\documentclass{article}
\usepackage[ruled,vlined]{algorithm2e}
\usepackage{booktabs}
\usepackage{multirow}
\usepackage{float}  
\usepackage{caption}
\usepackage{subcaption}
\usepackage{graphicx}
\usepackage{amsfonts}       
\usepackage{amsmath}        

\usepackage[preprint]{n_2025}
\usepackage{wrapfig} 
\usepackage{enumitem}

\usepackage[utf8]{inputenc} 
\usepackage[T1]{fontenc}    
\usepackage{hyperref}       
\usepackage{url}            
\usepackage{booktabs}       
\usepackage{amsfonts}       
\usepackage{nicefrac}       
\usepackage{microtype}      
\usepackage{xcolor}         
\usepackage{tabularx}
\usepackage{array}      

\title{HSPAG: A Unified and Data-efficient Molecular Property Modeling Framework via Multi-Level Contrastive Learning}

\author{%
  Ziyu Fan \\
  School of Computer Science and Engineering\\
  Central South University\\
  Changsha 410083, China \\
  \texttt{fzychina@csu.edu.cn} \\
  \And
  Zhijian Huang \\
  School of Computer Science and Engineering\\
  Central South University\\
  Changsha 410083, China \\
  \And
  Yahan Li \\
  School of Computer Science and Engineering\\
  Central South University\\
  Changsha 410083, China \\
  \And
  Siyuan Shen \\
  School of Computer Science and Engineering\\
  Central South University\\
  Changsha 410083, China \\
  \And
  Yunliang Wang \\
  School of Computer Science and Engineering\\
  Central South University\\
  Changsha 410083, China \\
  \And
  Zeyu Zhong \\
  School of Computer Science and Engineering\\
  Central South University\\
  Changsha 410083, China \\
  \And
  Shuhong Liu \\
  School of Computer Science and Engineering\\
  Central South University\\
  Changsha 410083, China \\
  \And
  Shuning Yang \\
  School of Computer Science and Engineering\\
  Central South University\\
  Changsha 410083, China \\
  \And
  Shangqian Wu \\
  School of Computer Science and Engineering\\
  Central South University\\
  Changsha 410083, China \\
  \And
  Min Wu\thanks{These authors are corresponding authors.} \\
  Institute for Infocomm Research\\
  Agency for Science, Technology and Research (A* STAR)\\
  Singapore 138632, Singapore\\
  \texttt{wumin@i2r.a-star.edu.sg} \\
  \And
  Lei Deng \footnotemark[1] \\
  School of Computer Science and Engineering\\
  Central South University\\
  Changsha 410083, China \\
  \texttt{leideng@csu.edu.cn} \\
}

\begin{document}

\maketitle
\clearpage

\appendix
\section{Dataset}
\label{A:dataset}
Our model HSPAG is trained on the ChEMBL 24 dataset~\cite{zhu2023pharmacophore}. The dataset comprises molecules built from 13 distinct atom types ($T = 13$), including H, B, C, N, O, F, Si, P, S, Cl, Se, Br, and I. Bond types are categorized into five groups ($R = 5$): non-bond, single, double, triple, and aromatic. All molecules are encoded using the SMILES. Each SMILES is then tokenized into a sequence based on atom symbols and structural delimiters. To ensure robust generalization and avoid data leakage, the dataset is split into training, validation, and test sets based on scaffolds. This scaffold-based split groups molecules by their core structures, preventing structurally similar compounds from appearing across different subsets. Both the validation and test sets consist of 2{,}000 molecules each, while the remaining data is used for training.

\section{Details of the HSPAG Framework}
\label{appendixMethod}
\subsection{Construction of the Example and Challenging Sets}
\label{appendix:data111}

To construct the ``example set'', we first perform scaffold-based partitioning on the full dataset, identifying 467,580 unique molecular scaffolds. Pairwise Levenshtein distances between scaffolds are computed, and those with a minimum distance of at least 3 are selected, resulting in 222,269 distinct scaffolds. For each scaffold, we randomly sample 1–2 representative molecules. Molecules for which ADMETlab 3.0~\cite{fu2024admetlab} fails to generate properties and molecules in valid or test dataset are removed. After filtering, the training set containing 236,920 molecules.

\begin{figure}[htbp]
  \centering
  \begin{minipage}[t]{0.2\textwidth}\vspace{0pt}%
    \centering
    \vspace{3pt} 
    \includegraphics[width=\linewidth]{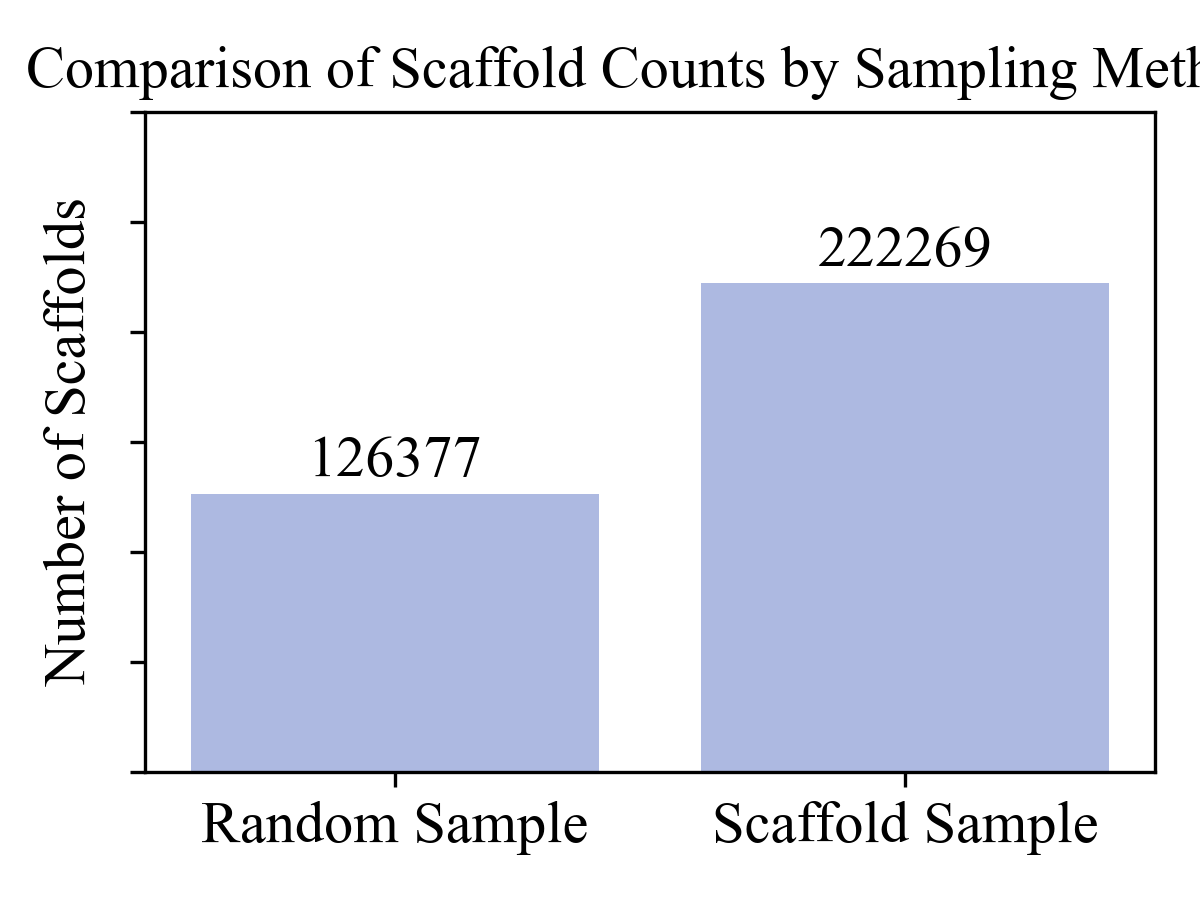}
    \caption{The number of scaffolds with different sampling strategy.}
    \label{fig:sca}
  \end{minipage}
  \begin{minipage}[t]{0.78\textwidth}\vspace{0pt}%
    \centering
    \includegraphics[width=\linewidth]{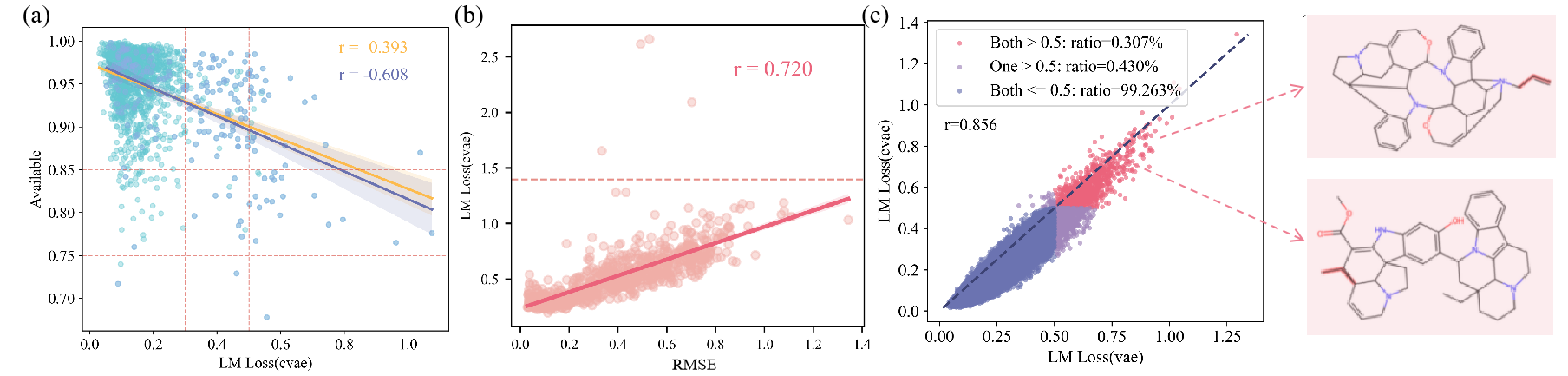}
    \caption{(a) Correlation of CVAE LM loss and Availability. (b) Correlation of RMSE and CVAE LM loss. (c) Correlation of CVAE LM loss and VAE LM loss.}
    \vspace{-0.6em} 
    \label{fig:active_learn}
  \end{minipage}
  \hfill
\end{figure}

\begin{wrapfigure}[21]{r}{0.45\textwidth} 
\vspace{-10pt}
  \centering
  \includegraphics[width=1\linewidth]{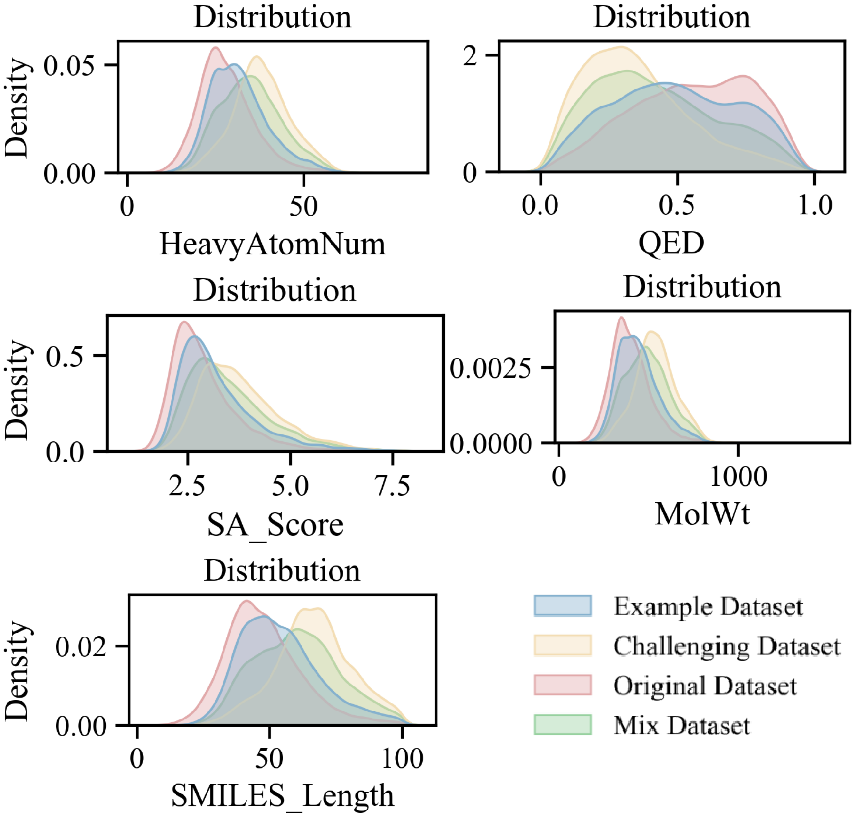}
  \caption{Distribution comparison across five molecular properties (HeavyAtomNum, QED, SA\_Score, MolWt, and SMILES\_Length) for different dataset splits.}
  \label{fenbubuu}
\end{wrapfigure}

The difference in the number of unique scaffolds between the randomly sampled example set and the scaffold-based sampled set from the training data is shown in the Fig.~\ref{fig:sca}. After training on the example set, as shown in Fig \ref{fig:active_learn}, we observe a negative correlation between the LM loss of CVAE and molecular generation performance (e.g., Availability), which aligns with intuition: molecules that are harder to reconstruct typically yield worse conditional generation results. This suggests that reconstruction loss can serve as a proxy for selecting difficult examples. However, since CVAE requires property conditions to compute this loss, it cannot be used directly on unlabeled data. To overcome this, we introduce an VAE without condition vectors. The VAE's reconstruction loss exhibits strong correlation with that of the CVAE (though slightly higher on average), making it a reliable alternative. Based on the VAE LM loss, we select generation-hard molecules as ``challenging set'' to supplement the example set—an approach analogous to uncertainty-based sampling in active learning.

To construct the ``challenging set'', we select molecules whose LM loss exceeds a threshold of 0.25, yielding 165,832 candidates. Among them, 43,067 molecules overlap with those in the training, validation, or test sets and are removed. The remaining 122,765 molecules correspond to 45,449 unique scaffolds, from which we again randomly sample 1–2 molecules per scaffold. After excluding samples without valid ADMETlab 3.0 outputs, we obtain a final set of 61,382 molecules to form the ``challenging set''.

As shown in Fig.~\ref{fenbubuu}, the example set closely aligns with the Original Dataset in all dimensions, demonstrating representativeness and full coverage of the major data distribution. In contrast, the challenging set shows a clear distribution shift, including higher molecular weights, lower synthetic accessibility scores, and longer SMILES—indicating greater structural complexity and modeling difficulty. Although the mix set integrates both example and challenging samples, it still exhibits a distribution deviation from the original dataset. 

\subsection{Learnable Logit Scaling in InfoNCE Loss}
\label{appendix:InfoNCE}
Following \textsc{CLIP}, we parameterize the inverse temperature \( 1/\tau \) using a learnable scalar logit\_scale, and apply it as a multiplicative factor to the similarity logits prior to the softmax:
\begin{equation}
\text{logits}_{i,j} = \exp(\text{logit\_scale}) \cdot \text{sim}(\mathbf{z}_{S_i}, \mathbf{z}_{P_j})
\end{equation}
To ensure numerical stability during training, we constrain \(\text{logit\_scale}\) such that its exponential does not exceed 100:
\begin{equation}
\tau = \frac{1}{\exp(\text{logit\_scale})}, \quad \text{with} \quad \text{logit\_scale} \leq \log 100
\end{equation}

\subsection{Variational Lower Bound}
~\label{appendix:cvae-details}
The training objective of the conditional variational autoencoder is to maximize the conditional log-likelihood of the observed sequence $S$ given a condition $P$, i.e., $\log p(S \mid P)$. Since this quantity is generally intractable, previous studies optimize the following variational lower bound (ELBO) instead\cite{zhu2023pharmacophore}:
\begin{equation}
\log p(S \mid P) \geq 
\mathbb{E}_{q_\phi(\mathbf{z} \mid S, P)} \left[
\log p_\theta(S \mid \mathbf{z}, P)
\right]
- D_{\mathrm{KL}} \left(
q_\phi(\mathbf{z} \mid S, P) \,\|\, p(\mathbf{z})
\right)
\end{equation}
Here, $q_\phi(\mathbf{z} \mid S, P)$ is an approximate posterior, and $p_\theta(S \mid \mathbf{z}, P)$ is the decoder likelihood conditioned on the latent variable $\mathbf{z}$ and input condition $P$. The latent variable $\mathbf{z} \in \mathbb{R}^d$ is drawn from a prior distribution $p(\mathbf{z}) = \mathcal{N}(\mathbf{0}, \mathbf{I})$.

The first term encourages the model to reconstruct the input sequence $S$ from the latent representation and condition, while the second term regularizes the approximate posterior to remain close to the prior. During training, the reparameterization trick is applied to enable gradient-based optimization:
\begin{equation}
\mathbf{z} = \boldsymbol{\mu} + \boldsymbol{\sigma} \odot \boldsymbol{\epsilon}, \quad 
\boldsymbol{\epsilon} \sim \mathcal{N}(\mathbf{0}, \mathbf{I})
\end{equation}
where $\boldsymbol{\mu}, \log \boldsymbol{\sigma}^2$ are outputs of neural networks conditioned on $(S, P)$.

This variational formulation allows the model to learn a smooth and expressive latent space while capturing the conditional structure in the data.

\subsection{Limitations}
\label{limittt}
\paragraph{Simulated property annotation.} In this study, molecular properties are primarily obtained through computational simulation or machine learning-based prediction, rather than experimental measurement~\cite{wu2024leveraging}. We employ ADMETlab 3.0~\cite{fu2024admetlab}, a comprehensive platform for evaluating ADMET (absorption, distribution, metabolism, excretion, and toxicity) properties of small molecules. The platform integrates multiple machine learning models trained on experimental data and predicts a wide range of physicochemical, pharmacokinetic, and toxicity endpoints. Many recent studies have adopted ADMETlab for large-scale property prediction~\cite{zhu2023pharmacophore,myung2024deep,wu2024leveraging,wu2022knowledge}. In addition, we complement structural property annotations using cheminformatics tools from RDKit to extract key molecular descriptors related to structure.

This large-scale annotation strategy facilitates the construction of high-quality, property-rich datasets, which are essential for training and evaluating deep molecular generative models. Although simulated properties may introduce some level of noise due to model uncertainty, they drastically reduce the cost and time associated with experimental data collection, and enable broader applicability across underexplored chemical spaces. With ongoing advances in predictive accuracy and coverage, such computational tools are expected to play an increasingly critical role in scalable, data-driven drug discovery\cite{wu2024leveraging}.

\paragraph{Missing structural constraints in ligand-based generation.} Although our model adopts a ligand-based generation strategy, which enables molecular design in the absence of protein structural information, it inevitably overlooks the three-dimensional geometry and physicochemical characteristics of the binding site. Compared to receptor-based generation methods, this limitation makes it difficult to ensure the spatial complementarity and binding affinity between the generated molecules and the target. Consequently, the generated compounds may exhibit suboptimal bioactivity or lack target specificity in real-world applications.

\subsection{Molecular generation and Property Prediction Pipeline}

\begin{figure}[htbp]
  \centering
  \begin{minipage}[t]{0.45\textwidth}\vspace{0pt}%
    \centering
    \includegraphics[width=\linewidth]{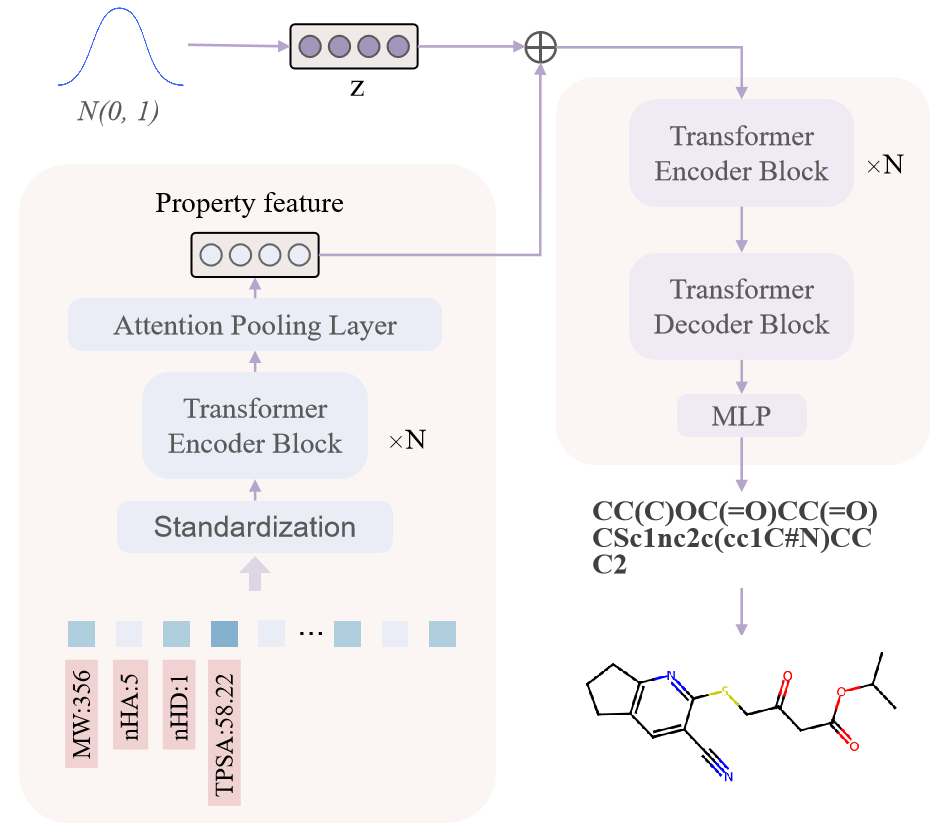}
    \caption{Molecular generation pipeline.}
    \vspace{-0.6em} 
    \label{xingzhi_gene}
  \end{minipage}
  \hfill
  \begin{minipage}[t]{0.54\textwidth}
 \vspace{8pt}%
    \centering
     \resizebox{0.8\linewidth}{!}{
    \includegraphics[width=\linewidth]{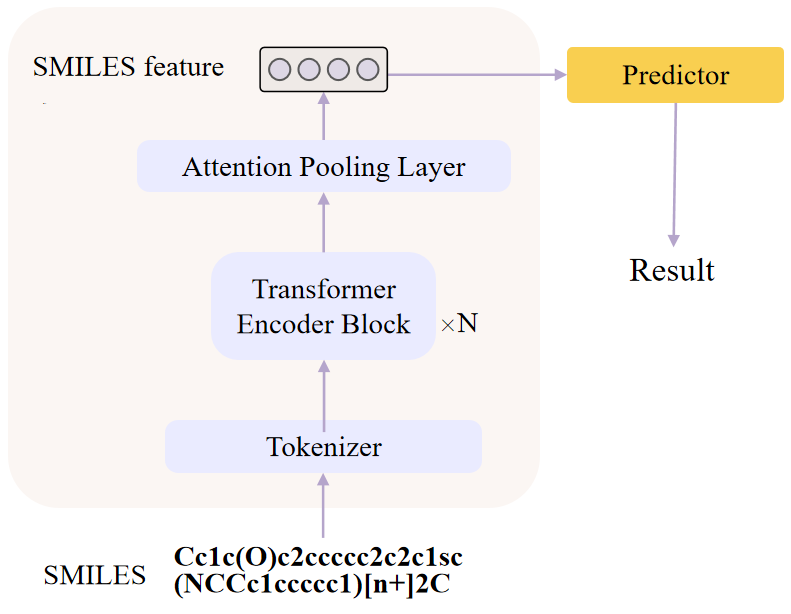}}
    \caption{Molecular property prediction Pipeline.}
    \label{xingzhi_pred}
  \end{minipage}
\end{figure}

The overall pipeline for molecular generation and property prediction is illustrated in Fig.~\ref{xingzhi_gene} and \ref{xingzhi_pred}. The fine-tuning process of property prediction does not freeze model parameters.

\section{Training details and model parameter settings}
\label{TDTAILS}
Our model adopts a Transformer-based architecture with a hidden dimension of 384. Both the encoder and decoder consist of 8 stacked Transformer blocks, each employing 8-head multi-head attention. The feed-forward network within each block has a dimensionality of 1024.

We use the Adam optimizer with a learning rate of $5 \times 10^{-5}$ and a weight decay of $1 \times 10^{-6}$. Gradient clipping is applied with a maximum gradient norm of 5. All experiments are conducted on NVIDIA A6000 GPU.

\section{Compared Baselines}
\label{appendix:tonghang}
This section outlines the baseline models used for comparison. All baselines are reproduced from their official open-source implementations under consistent experimental settings. The compared methods include:
\begin{itemize}[leftmargin=*, itemindent=0pt, itemsep=0.1em, parsep=0pt, nosep]
  \item LIMO~\cite{eckmann2022limo}, a VAE-based model leveraging a variational autoencoder generated latent space.
  \item CHEMFORMER~\cite{irwin2022chemformer}, a pre-trained molecular language model operating on SMILES representations.
  \item CProMG~\cite{li2023cpromg}, a protein-oriented controllable molecular generation framework that integrates hierarchical protein representations to generate novel molecules. We adapt this framework by replacing the protein input with property vectors.
  \item SPMM~\cite{chang2024bidirectional}, a scalable property-controlled molecular generation framework that supports multi-property conditioning by embedding discrete property prompts into the generation process.
  \item MD-VAE~\cite{10095212}, a multi-decoder VAE architecture that shares a single encoder while sampling distinct latent variables and applying a collaborative loss.
  \item AAE~\cite{makhzani2015adversarial}, a generative model that regularizes the latent space via a discriminator, enabling the learning of structured molecular representations and property-controllable molecule generation.
  \item MolMVC~\cite{huang2024molmvc}, a multi-view contrastive model for unified molecular representation across drug-related tasks.
  \item MolBART~\cite{chilingaryan2023molbart}, a self-supervised masked language model that achieves strong performance on molecular tasks while implicitly capturing chemistry-relevant features.
  \item BartSmiles~\cite{chilingaryan2024bartsmiles}, a self-supervised molecular language model that achieves leading results across a range of molecular tasks and implicitly captures key chemical substructures.
\end{itemize}

\section{HSPAG Aligns Closely with Real-World Molecular Distributions}
\label{appendix:noc_gene}
\subsection{Molecular generation process}

As show in Fig.~\ref{xingzhi_gene}, after training is complete, the CVAE generates molecular structures by sampling latent vectors from the learned prior distribution, typically a standard Gaussian. Conditioned on a given property vector, the decoder then generates the molecular sequence token by token in an autoregressive manner.

At each decoding step, the model predicts the probability distribution over possible next tokens based on the latent vector, the input property vector, and the tokens generated so far. This process continues until a special end-of-sequence token is produced, indicating the completion of the molecular representation. By leveraging the latent space and conditioning information, the model is capable of generating diverse molecules that satisfy the specified properties. This enables controllable and efficient exploration of the chemical space, which is valuable for tasks such as drug discovery and molecular optimization.

\subsection{Evaluation Metrics}
We evaluate molecular generation quality using the following standard metrics:

\begin{itemize}[leftmargin=*, itemindent=0pt, itemsep=0.1em, parsep=0pt, nosep]
    \item \textbf{Validity}: The percentage of generated SMILES that correspond to chemically valid molecules, as determined by RDKit.
    
    \item \textbf{Uniqueness}: The proportion of valid molecules that are unique within the generated set:
    \[
    \text{Uniqueness} = \frac{\# \text{Unique Valid SMILES}}{\# \text{Valid SMILES}}
    \]
    
    \item \textbf{Novelty}: The fraction of unique valid molecules that do not appear in the training set:
    \[
    \text{Novelty} = \frac{\# \text{Novel Molecules}}{\# \text{Unique Valid SMILES}}
    \]
    
    \item \textbf{Availability}: The percentage of generated molecules that are valid, unique and do not appear in the training set. 
    
    \item \textbf{SNN} (Similarity to Nearest Neighbor): The average Tanimoto similarity between each generated molecule and its nearest neighbor in the training set:
    \[
    \text{SNN} = \frac{1}{N} \sum_{i=1}^N \max_{j \in \text{train}} \text{Tanimoto}(f_i, f_j)
    \]
    
    \item \textbf{Frag} (Fragment Similarity): The cosine similarity between the fragment distribution of the generated set and that of the training set, using BRICS fragments.
    
    \item \textbf{FCD} (Fréchet ChemNet Distance): A distributional distance between generated and training molecules, computed using the Fréchet distance in the ChemNet embedding space:
    \[
    \text{FCD} = \left\|\mu_r - \mu_g\right\|_2^2 + \mathrm{Tr}\left(\Sigma_r + \Sigma_g - 2(\Sigma_r \Sigma_g)^{1/2} \right)
    \]
    where \( \mu_r, \Sigma_r \) and \( \mu_g, \Sigma_g \) are the means and covariances of real and generated ChemNet features, respectively.
    
    \item \textbf{IntDiv} (Internal Diversity): The average pairwise dissimilarity between generated molecules:
    \[
    \text{IntDiv} = 1 - \frac{2}{N(N-1)} \sum_{i < j} \text{Tanimoto}(f_i, f_j)
    \]
    where \( f_i \) and \( f_j \) are the fingerprint vectors of generated molecules.
    \item \textbf{Scaffold}: The number of scaffolds in the generated molecules.
    \item \textbf{nRMSE}: (Normalized Root Mean Square Error): The RMSE between the standardized input property conditions and the standardized properties of the generated molecules:
\[
\text{nRMSE} = \sqrt{ \frac{1}{K N} \sum_{k=1}^{K} \sum_{i=1}^{N} \left( \frac{ \hat{p}_i^{(k)} - p_i^{(k)} }{ \sigma^{(k)} } \right)^2 }
\]
where \( K \) is the number of property dimensions, \( N \) is the number of generated molecules, \( \hat{p}_i^{(k)} \) is the \( k \)-th predicted property of the \( i \)-th molecule, \( p_i^{(k)} \) is the corresponding input condition, and \( \sigma^{(k)} \) is the standard deviation of the \( k \)-th property over the training data.

\end{itemize}

\section{HSPAG Enables Accurate Property-Controlled Molecular Generation}
\label{appendix:c_gene}
\subsection{Settings and Metrics}
To systematically evaluate conditional generation performance, we select 1{,}000 target molecules whose scaffolds do not appear in the example set and challenging set. For each target, we extract its property vector as the condition and generate 2{,}000 SMILES.

In addition to standard metrics such as Validity, Uniqueness, and Novelty, we introduce a quantitative metric to assess condition-property alignment: normalized RMSE. Specifically, we compute the root mean square error between the properties of each generated molecule and the input condition, and normalize it using the mean and standard deviation of the corresponding property in the training set. The final score is averaged over all controlled properties. Due to the complexity of the calculation process, we used RDKit to calculate nRMSE based on the molecular features that can be obtained.

To further assess structural diversity, we count the number of unique scaffolds among the generated samples per condition, which reflects the model's generalization and exploration capacity in chemical space. Scaffold diversity is particularly valuable in drug discovery applications, as it facilitates scaffold hopping—the ability to generate novel core structures that retain desired bioactivity, thereby expanding the pool of viable lead compounds.

\subsection{Decoding Strategies for Generation}

In conditional molecular generation, the choice of decoding strategy significantly affects the diversity, validity, and controllability of generated samples. We consider three commonly used decoding methods: Top-$k$ sampling and Top-$p$ sampling~\cite{chang2024bidirectional,li2023cpromg}.

\begin{wraptable}[13]{r}{0.55\textwidth}
\vspace{-12pt}
  \centering
  \small
  \caption{Comparison of Top-$k$ and Top-$p$ sampling strategies under different property perturbation ratios.}
  \label{tab:ksample-psample}
  \renewcommand{\arraystretch}{1}
  \resizebox{0.99\linewidth}{!}{
  \begin{tabular}{lcccccccc}
\toprule
Method & Keep\_ratio & Val.$\uparrow$ & Uni.$\uparrow$ & Nov.$\uparrow$ & Avail.$\uparrow$  & RMSE & Scaf. \\
\midrule
K=1  & 1.0 & 0.9280 & 0.7167 & 0.9985 & 0.6640 & 0.2565 & 502.31 \\
K=1  & 0.9 & 0.9169 & 0.7708 & 0.9992 & 0.7061 & 0.2707 & 540.52 \\
K=2  & 1.0 & 0.9055 & 0.8653 & 0.9998 & 0.7825 & 0.2630 & 629.38 \\
K=2  & 0.9 & 0.8961 & 0.9069 & 0.9993 & 0.8121 & 0.2841 & 667.34 \\
K=3  & 1.0 & 0.9005 & 0.8918 & 0.9991 & 0.8023 & 0.2800 & 663.39 \\
K=3  & 0.9 & 0.8866 & 0.9197 & 0.9994 & 0.8149 & 0.2961 & 679.58 \\
P=90 & 1.0 & 0.9230 & 0.7954 & 0.9977 & 0.7324 & 0.2603 & 577.29 \\
P=90 & 0.9 & 0.9131 & 0.8434 & 0.9991 & 0.7694 & 0.2769 & 615.27 \\
P=80 & 1.0 & 0.9275 & 0.7680 & 0.9976 & 0.7106 & 0.2551 & 548.64 \\
P=80 & 0.9 & 0.9177 & 0.8211 & 0.9992 & 0.7529 & 0.2894 & 589.19 \\
P=70 & 1.0 & 0.9303 & 0.7379 & 0.9987 & 0.6855 & 0.2547 & 521.33 \\
P=70 & 0.9 & 0.9151 & 0.7883 & 0.9982 & 0.7207 & 0.2903 & 559.01 \\
\bottomrule
\end{tabular}}
\end{wraptable}

\textbf{Top-$k$ Sampling.}
Top-$k$ sampling limits the token selection space at each generation step to the $k$ most probable tokens, where $k$ is a predefined threshold. The next token is then sampled from this restricted subset according to the normalized probability distribution. This strategy eliminates low-probability tokens while retaining stochasticity to encourage molecular diversity. Larger $k$ values tend to increase diversity at the expense of validity and constraint satisfaction.

\textbf{Top-$p$ Sampling.}
Top-$p$ sampling dynamically selects the smallest set of tokens whose cumulative probability exceeds a threshold $p$. Unlike Top-$k$, which uses a fixed-size cutoff, Top-$p$ adjusts the candidate set based on the shape of the distribution, making it more adaptive. This approach retains a balance between diversity and coherence, and has shown success in generating fluent and chemically valid molecules, especially in settings with uncertain or perturbed conditions.

As shown in Tab.~\ref{tab:ksample-psample}, combined with input perturbation (i.e., the keep\_ratio controlling propertie retention), we perform ablation experiments under various $k$ and $p$ settings to assess the trade-offs between chemical validity, uniqueness, normalized RMSE, and scaffold diversity.

\section{HSPAG Supports Molecular Generation for Distributional Edge Properties}
\label{appendix:c_gene_d}

For each property range, we randomly sample 100 molecules from the dataset that satisfy the target condition and extract their property vectors as inputs. The model is then tasked with generating 1000 SMILES per condition vector.

Success (\%): percent of generated molecules within the target range. Diversity: One minus the average pairwise Tanimoto similarity between Morgan fingerprints.

Note that we omit the condition \(500 \leq \text{MW} \leq 550\) used in prior work due to its relatively sufficient coverage in our training data.

\section{\textsc{HSPAG} Enables Accurate Prediction of Molecular Properties}
\label{appendix:xingzhi}
\subsection{Settings}

To ensure robust evaluation, we adopt the scaffold-based split strategy with an 8:1:1 ratio for training, validation, and test sets, respectively, and report average results over three random seeds~\cite{fang2023knowledge}. This splitting scheme prevents scaffold leakage between training and test data, which better reflects real-world generalization.

To avoid information leakage during pretraining, we mask the property values of all molecules that appear in the downstream tasks, ensuring that their properties are never observed by the model during pretraining. This design guarantees fair and unbiased evaluation of the model's transfer performance.

\subsection{Structure–Property Association and Embedding Sensitivity Analysis}
\label{222222}
To further investigate the strength of \textsc{HSPAG}'s unsupervised molecular representation, we evaluate its ability to capture structure–property associations. Molecular substructures, such as aromatic rings or specific functional groups, often serve as key indicators of biochemical properties.

\begin{table}[H]
\centering

\caption{Cramér’s V between molecular substructure counts and downstream properties.}
\small
\resizebox{0.8\linewidth}{!}{%
\begin{tabular}{lccccccccc}

\toprule
Pre-training & BBBP & Tox21 & ToxCast  & ClinTox & MUV & HIV & Bace & ESOL & Lipophilicity \\
\midrule
allylic    & 0.1602 & 0.1345 & 0.1156 & 0.0935 & 0.0413 & 0.0280 & 0.1186 & 0.1092 & 0.0289 \\
amide      & 0.2692 & 0.0490 & 0.0858 & 0.1326 & 0.0235 & 0.0689 & 0.2556 & 0.1553 & 0.0699 \\
amidine    & 0.0360 & 0.0291 & 0.0142 & 0.0158 & 0.0117 & 0.0396 & 0.1328 & -      & 0.0296 \\
azo        & 0.0400 & 0.0399 & 0.0393 & 0.0123 & 0.0007 & 0.2082 & -      & 0.0309 & 0.0253 \\
benzene    & 0.1476 & 0.1632 & 0.1691 & 0.1112 & 0.0289 & 0.1374 & 0.1091 & 0.3439 & 0.1202 \\
epoxide    & 0.0273 & 0.0481 & 0.0449 & 0.0049 & 0.0005 & 0.0086 & -      & 0.0247 & 0.0477 \\
ether      & 0.2314 & 0.0694 & 0.1060 & 0.1023 & 0.0185 & 0.0498 & 0.1821 & 0.1093 & 0.0926 \\
furan      & 0.0635 & 0.0257 & 0.0387 & 0.0061 & 0.0311 & 0.0148 & 0.0135 & 0.0409 & 0.0254 \\
guanido    & 0.0765 & 0.0201 & 0.0509 & 0.0286 & 0.0057 & 0.0094 & 0.1088 & 0.0247 & 0.0790 \\
halogen    & 0.1488 & 0.0849 & 0.1827 & 0.0908 & 0.0143 & 0.0347 & 0.2353 & 0.2175 & 0.0874 \\
imidazole  & 0.0601 & 0.0427 & 0.0492 & 0.1212 & 0.0102 & 0.0398 & 0.1280 & 0.0187 & 0.0497 \\
imide      & 0.0951 & 0.0246 & 0.0401 & 0.0518 & 0.0094 & 0.0188 & -      & 0.1347 & 0.0216 \\
lactam     & 0.4263 & 0.0184 & 0.0116 & 0.0543 & 0.0006 & 0.0048 & -      & -      & 0.0179 \\
morpholine & 0.0512 & 0.0126 & 0.0343 & 0.0425 & 0.0068 & 0.0101 & 0.0668 & -      & 0.0507 \\
N\_O       & 0.0438 & 0.0195 & 0.0467 & 0.0709 & 0.0195 & 0.0144 & 0.0537 & 0.0288 & 0.0253 \\
oxazole    & 0.0126 & 0.0184 & 0.0321 & 0.0123 & 0.0079 & 0.0080 & 0.0364 & 0.0368 & 0.0485 \\
piperidine & 0.1450 & 0.0305 & 0.0844 & 0.0418 & 0.0079 & 0.0226 & 0.0935 & 0.0445 & 0.0709 \\
piperzine  & 0.0509 & 0.0214 & 0.0421 & 0.0648 & 0.0111 & 0.0192 & 0.0063 & 0.0309 & 0.0411 \\
pyridine   & 0.0598 & 0.0402 & 0.0549 & 0.0833 & 0.0129 & 0.0300 & 0.1747 & 0.0785 & 0.0686 \\
tetrazole  & 0.1161 & 0.0158 & 0.0251 & 0.0286 & 0.0083 & 0.0123 & 0.0334 & -      & 0.0234 \\
thiazole   & 0.1389 & 0.0521 & 0.0345 & 0.0183 & 0.0118 & 0.0173 & 0.0539 & 0.0137 & 0.0121 \\
thiophene  & 0.0356 & 0.0467 & 0.0472 & 0.0113 & 0.0166 & 0.0081 & 0.0438 & 0.0203 & 0.0359 \\
urea       & 0.0790 & 0.0236 & 0.0506 & 0.0268 & 0.0079 & 0.0329 & 0.0516 & 0.0722 & 0.0169 \\
\bottomrule
\end{tabular}}
\label{CCCCVV}
\end{table}

We follows the foundation laid by earlier studies~\cite{wang2023evaluating}, selected 23 representative substructures, categorized into three classes: rings (e.g., benzene, furan, thiophene), functional groups (e.g., amide, amidine, nitrile, urea), and redox-active motifs (e.g., allyl group). For each MoleculeNet dataset, we computed the Cramér’s V~\cite{wang2023evaluating} score between substructure presence and property labels to assess their statistical association. Cramér’s V is a chi-squared-based measure that quantifies the strength of dependence between two categorical variables:

\begin{equation}
V = \sqrt{\frac{\chi^2}{n \cdot \min(k - 1, r - 1)}}
\end{equation}

where $n$ is the sample size, $k$ and $r$ are the number of substructure types and property classes respectively, and $\chi^2$ is the chi-squared statistic defined as:

\begin{equation}
\chi^2 = \sum_{i,j} \frac{\left(n_{(i,j)} - \frac{n_{(i,\cdot)} \cdot n_{(\cdot,j)}}{n} \right)^2}{\frac{n_{(i,\cdot)} \cdot n_{(\cdot,j)}}{n}}
\end{equation}

Here, $n_{(i,j)}$ denotes the co-occurrence count of substructure $i$ and label $j$, while $n_{(i,\cdot)}$ and $n_{(\cdot,j)}$ represent marginal counts.

ESOL is classified into three categories: low ($\le -3$), medium (between $-3$ and $1$), and high ($\ge 1$). Lipophilicity is also divided into three categories: low ($\le -3$), medium (between $-3$ and $1$), and high ($\ge 1$).

Substructures with $V > 0.1$ were deemed property-relevant (Tab.~\ref{CCCCVV}). To assess how well pretrained models separate molecules with and without such structures, we extracted embeddings using the frozen encoders (i.e., without fine-tuning), and computed the Davies–Bouldin (DB) score between the two embedding clusters. A lower DB score indicates greater inter-cluster separability and thus better structure sensitivity in the learned representations.

Formally, the DB score is defined as:

\begin{equation}
\text{DB} = \frac{1}{K} \sum_{i=1}^{K} \max_{j \neq i} \left( \frac{\sigma_i + \sigma_j}{d(c_i, c_j)} \right)
\end{equation}

where $K=2$ for the binary grouping, $\sigma_i$ is the average distance from embeddings in cluster $i$ to its centroid $c_i$, and $d(c_i, c_j)$ is the Euclidean distance between cluster centroids.

Empirical results show that \textsc{HSPAG} consistently achieves the lowest average DB scores across all models, demonstrating its superior ability to encode substructure-sensitive features and preserve structure–property mappings during pretraining.

\begin{table}[ht]
\centering
\caption{Substructure-based comparison on BACE dataset.}
\scriptsize
\begin{tabular}{lccccccccccc}
\toprule
 & Amide & Halogen & Ether & Pyridine & Amidine & Imidazole & Piperidine & Allylic & Benzene & Guanido & Mean \\
\midrule
HSPAG     & 3.605 & 4.097 & 5.196 & 4.188 & 2.629 & 1.798 & 3.641 & 2.750 & 1.838 & 2.399 & 3.214 \\
MolMVC  & 4.017 & 4.819 & 5.014 & 4.741 & 3.487 & 2.339 & 4.707 & 2.549 & 1.824 & 3.667 & 3.716 \\
Spmm    & 3.949 & 4.451 & 6.456 & 4.237 & 2.573 & 2.380 & 3.625 & 3.449 & 2.531 & 2.644 & 3.630 \\
BartSmiles    & 3.425 & 7.389 &10.914 & 3.075 & 1.710 & 2.154 & 1.480 & 3.213 & 4.461 & 1.623 & 3.944 \\
\bottomrule
\end{tabular}
\end{table}

\begin{table}[H]
\centering
\begin{minipage}[t]{0.59\textwidth}\vspace{0pt}%
\vspace{-10pt} 
\centering
\caption{Substructure-based comparison on BBBP dataset.}
\vspace{6pt}
\label{tab:A3}
\renewcommand{\arraystretch}{1}
\resizebox{1.3\linewidth}{!}{ 
\begin{tabular}{lcccccccccc}
\toprule
 & Lactam & Amide & Ether & Allylic & Halogen & Benzene & Piperidine & Thiazole & Tetrazole & Mean \\
\midrule
HSPAG     & 2.284 & 5.581 & 5.022 & 2.835 & 5.060 & 3.778 & 3.899 & 2.909 & 2.176 & 3.727 \\
MolMVC  & 2.181 & 5.341 & 4.659 & 2.494 & 6.937 & 4.241 & 4.502 & 2.692 & 2.380 & 3.936 \\
Spmm    & 2.553 & 5.908 & 6.840 & 3.449 & 5.753 & 4.575 & 5.216 & 2.824 & 2.427 & 4.393 \\
BartSmiles    & 2.397 & 8.691 & 5.062 & 2.547 &10.426 & 5.468 & 7.138 & 2.751 & 2.139 & 5.180 \\
\bottomrule
\end{tabular}}
\end{minipage}
\hfill
\begin{minipage}[t]{0.38\textwidth}\vspace{0pt}%
\vspace{-10pt} 
\centering
\caption{Lipophilicity dataset.}
\vspace{6pt}
\label{tab:A3}
\renewcommand{\arraystretch}{1}
\resizebox{0.8\linewidth}{!}{ 
\hspace*{70pt} 
  \begin{tabular}{lc}
\toprule
 & Benzene  \\
\midrule
HSPAG     & 4.711 \\
MolMVC  & 4.734  \\
Spmm    & 5.242  \\
BartSmiles    & 9.156  \\
\bottomrule
\end{tabular}}
\end{minipage}
\end{table}

\begin{table}[ht]
\centering
\begin{minipage}[t]{0.3\textwidth}\vspace{0pt}%
\vspace{-10pt} 
\centering
\caption{Substructure-based comparison on Tox21 dataset.}
\vspace{6pt}
\label{tab:constrained-property-gen}
\renewcommand{\arraystretch}{0.9}
\resizebox{\textwidth}{!}{ 
\begin{tabular}{lccc}
\toprule
 & Benzene & Allylic & Mean \\
\midrule
HSPAG     & 3.698 & 3.537 & 3.618 \\
MolMVC  & 4.678 & 3.467 & 4.073 \\
Spmm    & 5.050 & 5.162 & 5.106 \\
BartSmiles    & 4.219 & 4.661 & 4.440 \\
\bottomrule
\end{tabular}}
\end{minipage}
\hfill
\begin{minipage}[t]{0.3\textwidth}\vspace{0pt}%
\vspace{-10pt} 
\centering
\caption{Substructure-based comparison on HIV dataset.}
\vspace{6pt}
\label{tab:constrained-property-gen}
\renewcommand{\arraystretch}{0.9}
\resizebox{\textwidth}{!}{ 
\begin{tabular}{lccc}
\toprule
 & Azo & Benzene & Mean \\
\midrule
HSPAG     & 3.892 & 4.586 & 4.239 \\
MolMVC  & 4.029 & 4.894 & 4.462 \\
Spmm    & 4.887 & 5.096 & 4.992 \\
BartSmiles    & 4.610 & 7.245 & 5.928 \\
\bottomrule
\end{tabular}}
\end{minipage}
\hfill
\begin{minipage}[t]{0.38\textwidth}\vspace{0pt}%
\vspace{-10pt} 
\centering
\caption{Substructure-based comparison on ToxCast dataset.}
\vspace{6pt}
\label{tab:A3}
\renewcommand{\arraystretch}{1}
\resizebox{\textwidth}{!}{ 
\begin{tabular}{lccccc}
\toprule
 & Halogen & Benzene & Allylic & Ether & Mean \\
\midrule
HSPAG     & 3.780 & 4.372 & 3.830 & 5.848 & 4.458 \\
MolMVC  & 5.334 & 4.722 & 3.505 & 4.715 & 4.569 \\
Spmm    & 5.347 & 5.197 & 5.289 & 5.982 & 5.454 \\
BartSmiles    & 9.081 & 4.684 & 5.066 & 6.608 & 6.360 \\
\bottomrule
\end{tabular}}
\end{minipage}
\end{table}

\begin{table}[H]
\centering
\begin{minipage}[t]{0.53\textwidth}\vspace{0pt}%
\vspace{-10pt} 
\centering
\caption{Substructure-based comparison on ESOL dataset.}
\vspace{6pt}
\label{tab:A3}
\renewcommand{\arraystretch}{1}
\resizebox{\textwidth}{!}{ 
\begin{tabular}{lccccccc}
\toprule
 & Benzene & Halogen & Amide & Imide & Ether & Allylic & Mean \\
\midrule
HSPAG     & 2.734 & 2.795 & 3.607 & 2.431 & 3.915 & 2.871 & 3.059 \\
MolMVC  & 4.118 & 3.555 & 5.014 & 2.241 & 4.626 & 3.183 & 3.790 \\
Spmm    & 4.069 & 3.945 & 3.246 & 2.638 & 4.613 & 4.104 & 3.769 \\
BartSmiles    & 3.026 & 6.197 & 3.369 & 2.760 & 6.288 & 3.542 & 4.197 \\
\bottomrule
\end{tabular}}
\end{minipage}
\hfill
\begin{minipage}[t]{0.42\textwidth}\vspace{0pt}%
\vspace{-10pt} 
\centering
\caption{Substructure-based comparison on ClinTox dataset.}
\vspace{6pt}
\label{tab:A3}
\renewcommand{\arraystretch}{1}
\resizebox{\textwidth}{!}{ 
\begin{tabular}{lccccc}
\toprule
 & Amide & Imidazole & Benzene & Ether & Mean \\
\midrule
HSPAG     & 4.634 & 3.655 & 4.488 & 4.606 & 4.346 \\
MolMVC  & 5.248 & 3.797 & 4.969 & 4.443 & 4.614 \\
Spmm    & 6.679 & 5.534 & 5.352 & 6.870 & 6.109 \\
BartSmiles    & 6.860 & 6.110 & 6.928 & 5.433 & 6.333 \\
\bottomrule
\end{tabular}}
\end{minipage}
\end{table}

\subsection{Property-Aware Embedding Reflects Structure-Property Correlations}

To further investigate whether \textsc{HSPAG} captures informative molecular properties in its latent representations, we design a nearest-neighbor correlation analysis on MoleculeNet datasets. Specifically, we randomly select 100 molecules with distinct scaffolds from each dataset as queries. For each query molecule, we compute its embedding using the pretrained model and retrieve its most similar counterpart (based on cosine similarity) from the remaining dataset, forming 100 molecular pairs.

\begin{wrapfigure}[14]{r}{0.4\textwidth} 
\vspace{-10pt}
  \centering
  \includegraphics[width=1\linewidth]{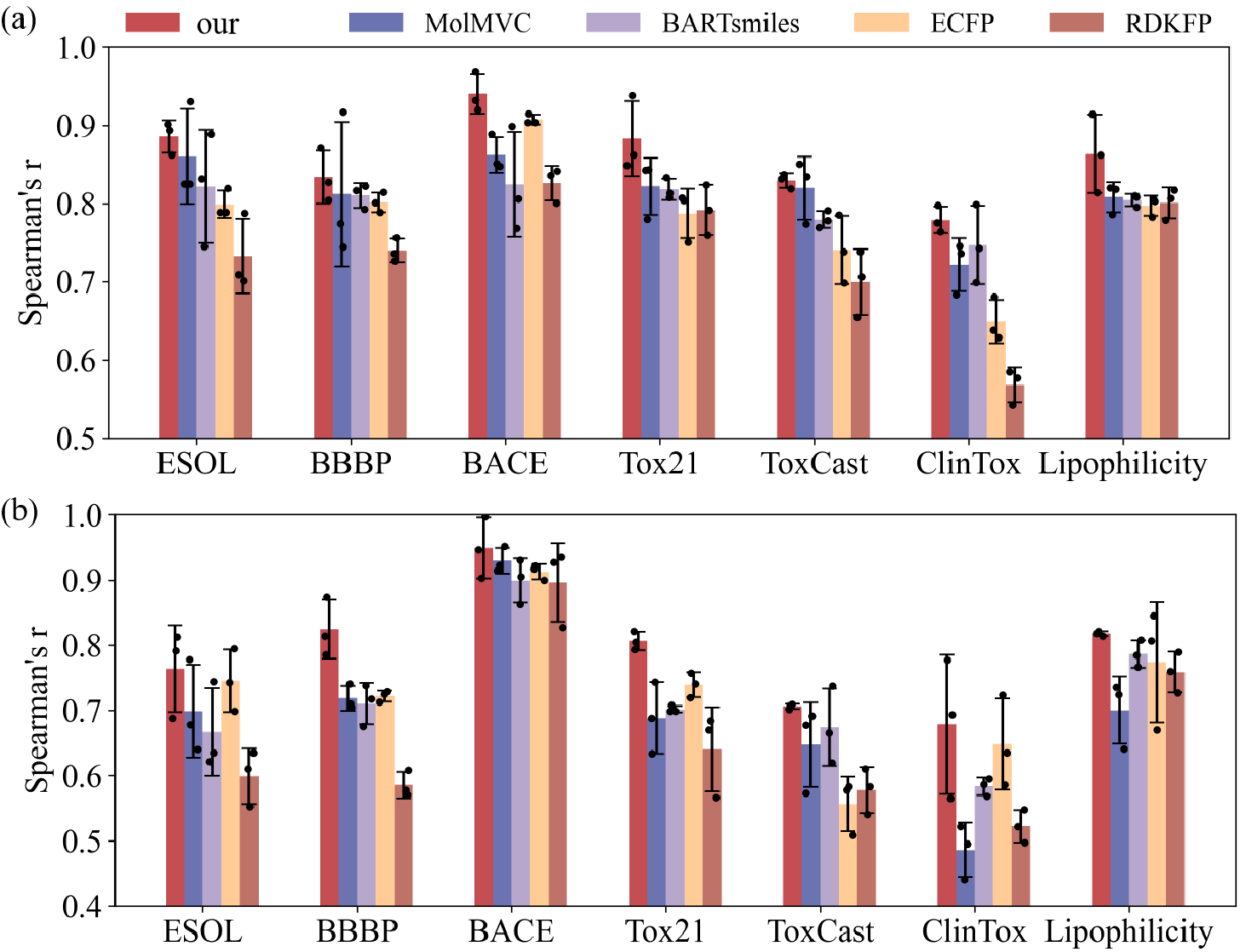}
  \caption{Spearman correlation of SAscore and QED.}
  \label{sim1}
\end{wrapfigure}

We evaluate the alignment between embeddings and chemical properties by calculating the Spearman correlation of QED scores between each query and its nearest neighbor. QED serves as a holistic metric combining multiple pharmaceutically relevant properties such as MW, LogP, TPSA, and HBA/HBD, offering a comprehensive view of molecule-level drug-likeness beyond individual properties.

To provide baseline comparisons, we repeat the same procedure using classical structural similarity measures: ECFP. Nearest neighbors are retrieved based on Tanimoto similarity in the respective fingerprint space, and QED correlation is similarly reported.

Additionally, to assess whether the learned embeddings preserve structural proximity, we compute structural similarity (using ECFP) between the retrieved molecular pairs. Since these fingerprint-based methods directly define structure similarity, we exclude ECFP and RDKFP from structure-only comparison.

\begin{wrapfigure}[15]{r}{0.8\textwidth} 
\vspace{-6pt}
  \centering
  \includegraphics[width=1\linewidth]{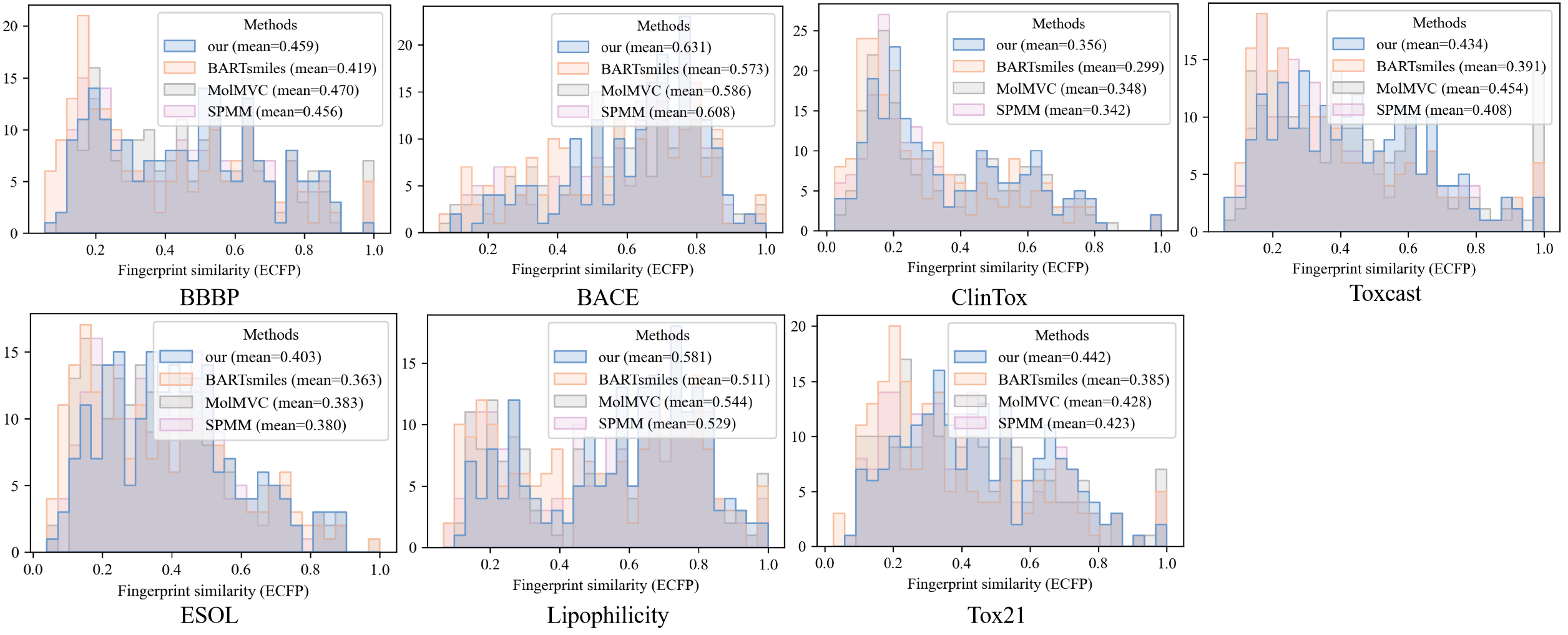}
  \caption{Structural similarity between the retrieved molecular pairs.}
  \label{sim2}
\end{wrapfigure}

Fig.~\ref{sim1} and \ref{sim2} indicate that \textsc{HSPAG} embeddings consistently yield higher QED correlation while maintaining strong structural alignment, demonstrating the model’s ability to jointly encode both structural and property-aware information in a unified representation space.

\subsection{Drug Repurposing via Virtual Screening} \label{casss}

\begin{figure}[h]
  \centering
  \includegraphics[width=0.99\linewidth]{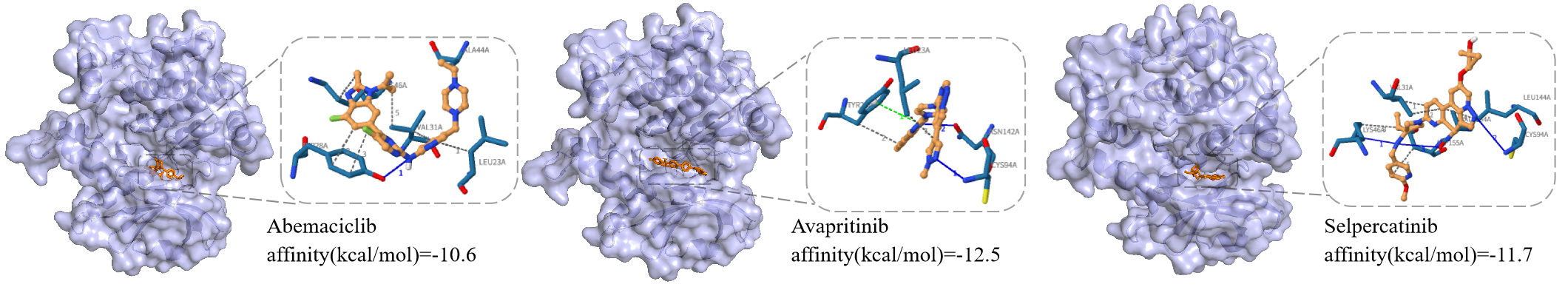}
  \caption{The protein-ligand structure (PDB ID: 7SIU46) was utilized as a reference for the identification of the binding pocket. The 3D structures of Abemaciclib, Selpercatinib, and Avapritinib were downloaded from PubChem. The interactions between molecules and HPK1 were profiled by PLIP.}
  \label{re2}
\end{figure}

Given the ability of our molecular representations to capture both structural and property-aware similarities, we further evaluate the applicability of \textsc{HSPAG} in real-world virtual screening and drug repurposing tasks. Specifically, we target Hematopoietic Progenitor Kinase 1 (HPK1), a well-established immuno-oncology target implicated in tumor immune suppression.

We obtain the data from \cite{li2023knowledge}, including a dataset containing 4,442 molecules with experimentally determined inhibitory activity against HPK1, measured by the negative logarithm of IC\textsubscript{50} values (pIC\textsubscript{50}). As the candidate pool, we compile a library of 2,580 FDA-approved drugs from DrugBank.

We identify highly active HPK1 inhibitors from the benchmark set with pIC\textsubscript{50} $\geq$ 7 and retrieve the top-10 most similar molecules from the FDA drug set for each active compound using the \textsc{HSPAG} embeddings. After aggregating and ranking all retrieved candidates, the top three frequently selected molecules are Abemaciclib, Selpercatinib, and Avapritinib. Notably, Abemaciclib has been experimentally validated in prior work as a potential HPK1 inhibitor, supporting the practical effectiveness of our retrieval approach~\cite{li2023knowledge}.

\section{HSPAG Demonstrates Excellent Molecular Editing Capability}
\label{appendix:bianji}
\subsection{Molecular Editing}
Molecular editing aims to perform targeted modifications on a given molecule such that specific properties are optimized, while retaining the molecule’s important structural characteristics. This task is critical in drug discovery scenarios such as lead optimization, where minimal and controlled changes are desired to fine-tune pharmacological profiles.

We propose a property-aware editing strategy that explicitly leverages inter-property correlations during conditional generation. 

Given a target molecule and a desired change in a specific property, we mask other attributes that exhibit high correlation with it. This avoids inconsistency among property dimensions that may otherwise mislead the model. For instance, increasing the molecular weight of a molecule typically correlates with a higher heavy atom count; failing to mask such correlated dimensions can result in conflicting inputs and suboptimal edits. A correlation threshold $\mu$ is used to control the masking scope: smaller $\mu$ leads to more dimensions being masked.

\begin{figure}[H]
  \centering
  \includegraphics[width=0.75\linewidth]{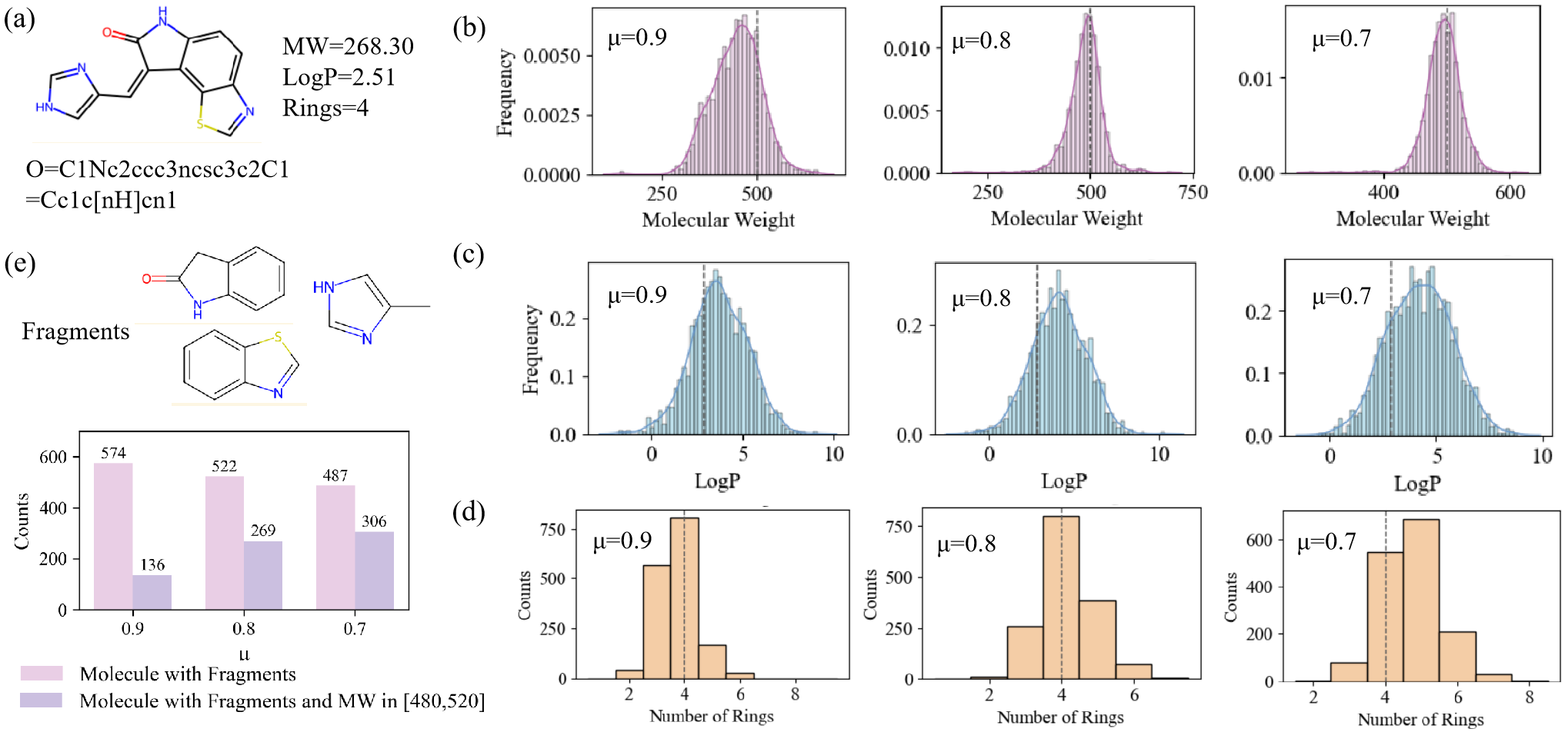}
  \caption{Molecular editing case (Molecule Weight = 500).}
  \label{edit1}
\end{figure}

\begin{figure}[H]
  \centering
  \includegraphics[width=0.75\linewidth]{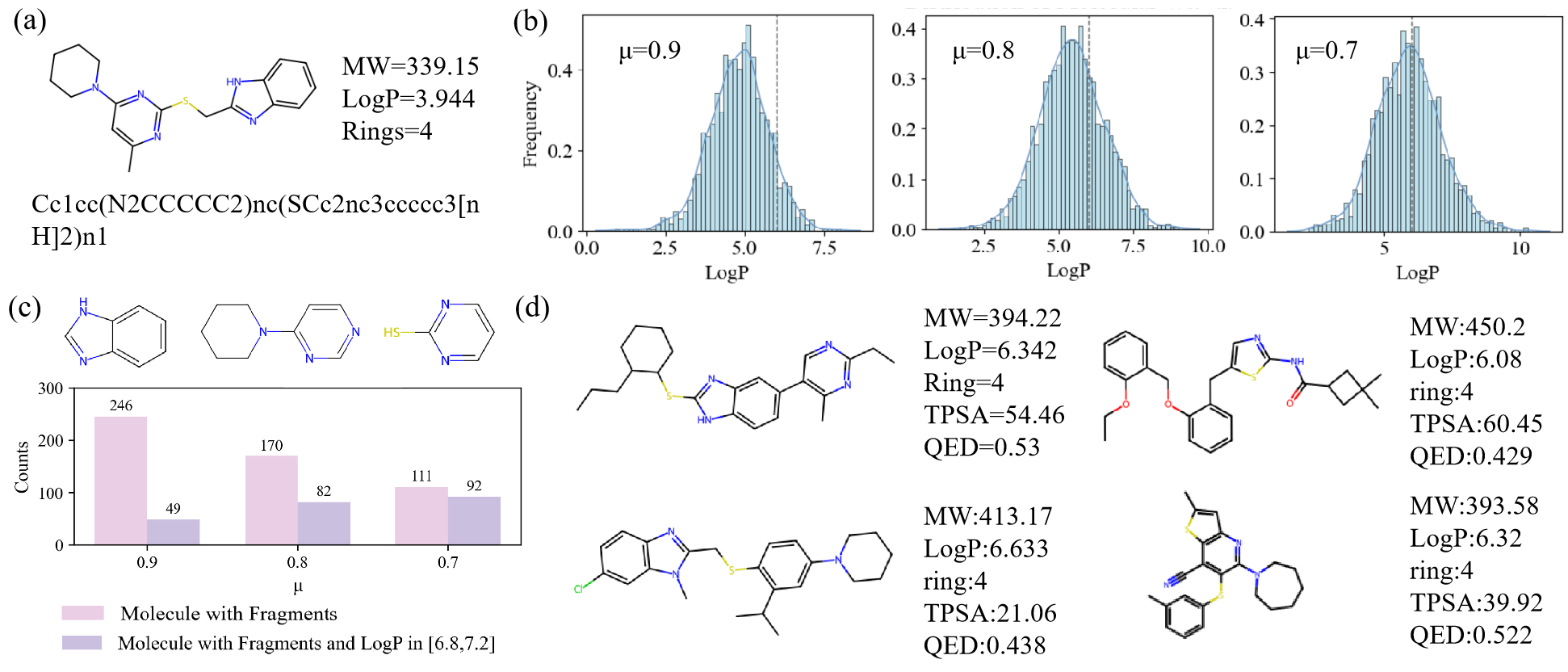}
  \caption{Molecular editing case (LogP = 6.5).}
  \label{edit2}
\end{figure}

To encourage diversity among generated molecules, we further apply random masking to additional non-target properties, introducing controlled uncertainty into the decoding process.

We conduct case studies on three representative editing tasks: increasing LogP to 6.0 (set 6.5), setting MW to 500 and setting QED to 0.6. For each task, we measure editing success by checking whether the generated molecule (i) satisfies the target property constraint and (ii) retains representative fragments (e.g., functional groups or scaffold motifs) extracted from the original molecule. If both conditions are met, the edit is considered successful.

\begin{wrapfigure}[10]{l}{0.5\textwidth} 
\vspace{-3pt}
  \centering
  \includegraphics[width=1\linewidth]{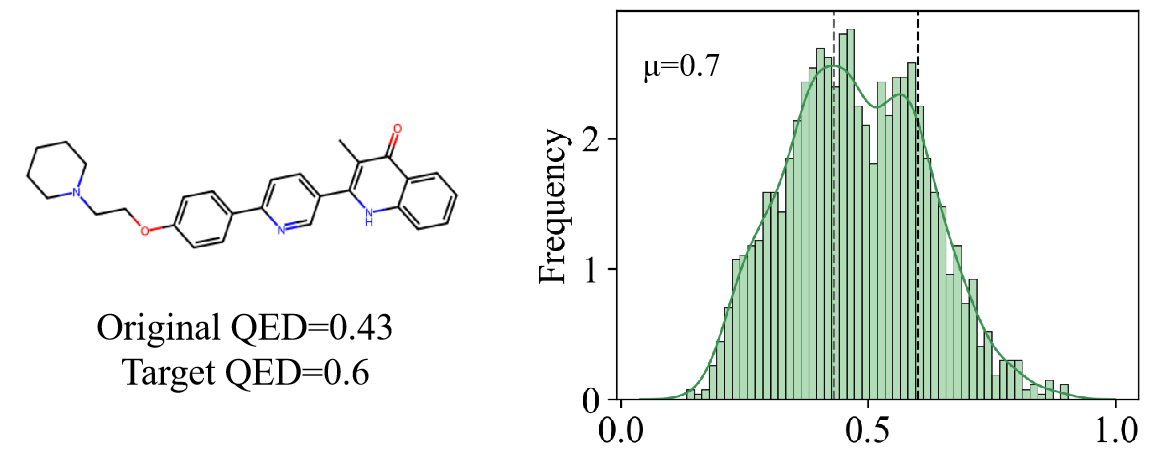}
  \caption{Molecular editing case (QED = 0.6).}
  \label{qed}
\end{wrapfigure}

As shown in Fig.~\ref{edit1} and Fig.~\ref{edit2}, experimental results indicate that decreasing the correlation threshold $\mu$ improves the property alignment rate but reduces structural consistency with the original molecule. This trade-off becomes more pronounced when the gap between the original and target properties is large, where the model may prefer full reconstruction over editing. In such cases, we also observe bimodal distributions in generated property values(Fig. \ref{qed}). Therefore, selecting an appropriate $\mu$ is crucial to balancing property accuracy and structural fidelity.

\subsection{Case Study: HSPAG is used for lead compound optimization}
\label{appendix:true_case}
To further validate the practical utility of our proposed molecular editing framework in real-world drug optimization scenarios, we conducted case studies on two compounds with clearly defined optimization objectives~\cite{xiong2021strategies}.

\begin{figure}[H]
  \centering
  \includegraphics[width=0.8\textwidth]{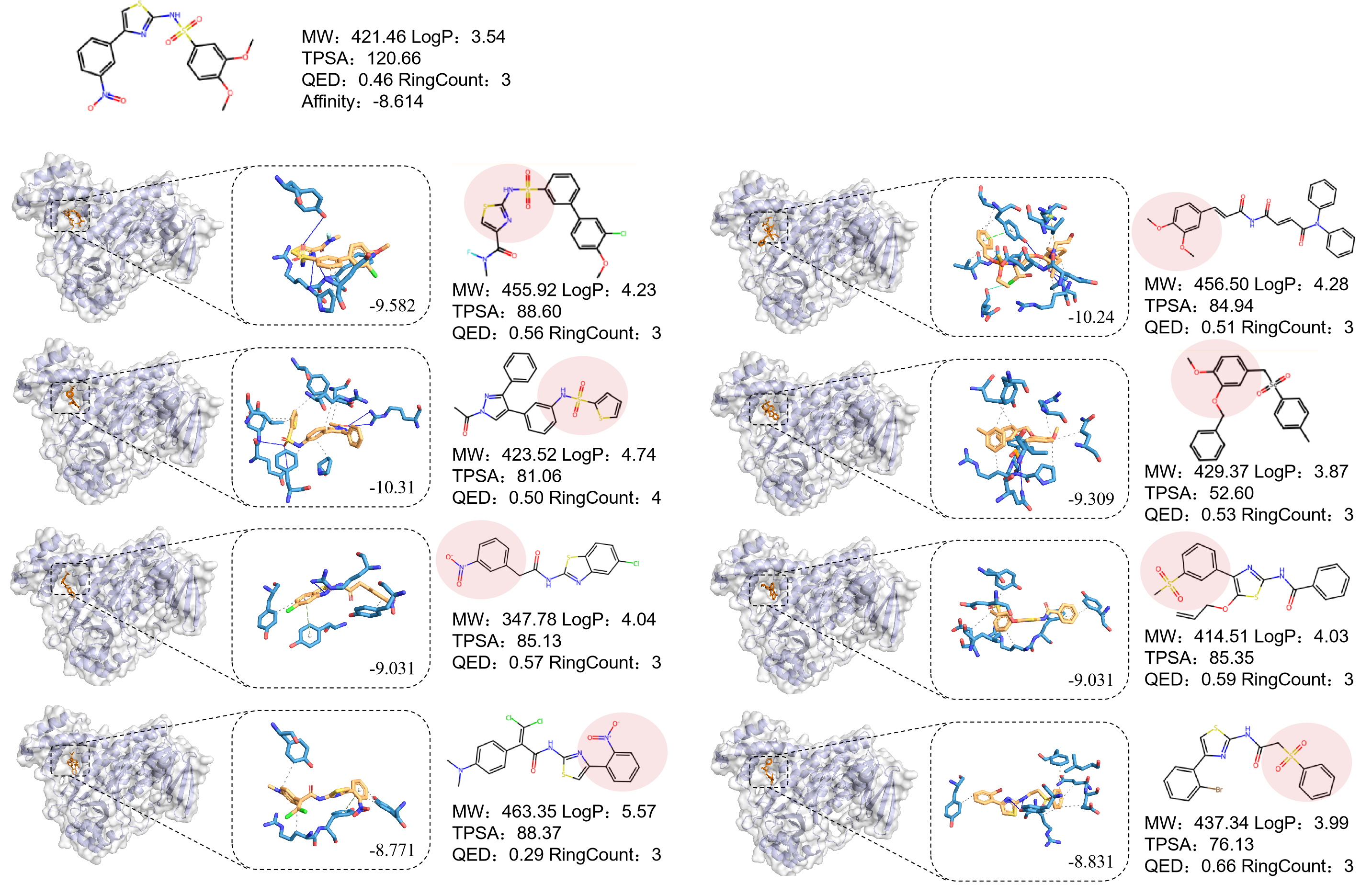}
  \caption{Structure optimization of Ro-61-8048 toward reduced tPSA.}
  \label{drug2}
\end{figure}

\begin{figure}[H]
  \centering
  \includegraphics[width=0.8\textwidth]{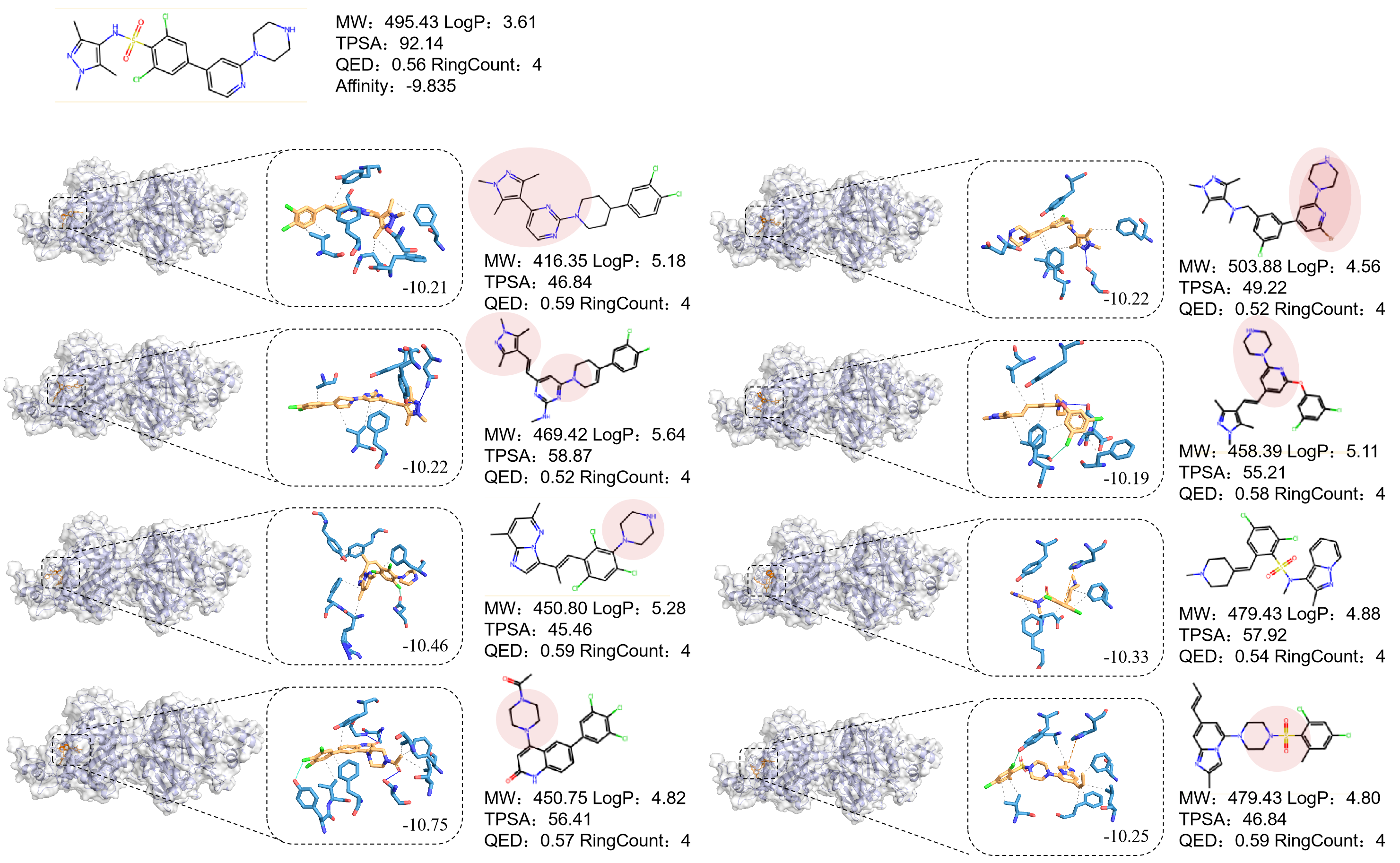}
  \caption{Structure optimization of DDD85646 toward reduced tPSA and enhancing logP.}
  \label{drug3}
\end{figure}

\textbf{Case 1 — KMO Inhibitor.}  
As shown in Fig~\ref{drug2}, the original molecule (Ro-61-8048) exhibits a high topological polar surface area (TPSA  $\approx 160$), which limits its BBB permeability. We set the target TPSA to 80 and masked strongly correlated properties (e.g., hydrogen bond donors/acceptors). This guides the model to lower polarity while preserving the core fragments (highlighted in red). All generated candidates were docked using AutoDock Vina against the crystal structure (PDB ID: 5Y66), and some showed lower binding free energies than the original molecule ($-8.614\ \text{kcal}\cdot\text{mol}^{-1}$), with TPSA values matching the specified goal. These results indicate an enhanced potential for brain exposure.

\textbf{Case 2 — NMT Inhibitor.}  
As shown in Fig~\ref{drug3}, the initial compound (DDD85646) has TPSA = 92 and $\text{logP} = 3.61$, which impairs passive diffusion. We set dual editing objectives: reduce TPSA to 42 and increase logP to 6. After masking highly correlated properties, our model generated molecules that retained the key fragments (red) and met both property constraints. All candidates achieved lower docking scores than the original molecule ($-9.835\ \text{kcal}\cdot\text{mol}^{-1}$), suggesting maintained or improved binding affinity.All generated candidates were docked using AutoDock Vina against the crystal structure (PDB ID: 3IWE).

These results demonstrate that \textsc{HSPAG} is capable of performing fine-grained molecular modifications under multi-property constraints. It not only satisfies specific quantitative property goals but also preserves essential structure-activity motifs, offering a powerful tool for lead optimization in molecular design.

\section{Ablation Results on Molecular Retrieval Task}

\begin{figure}[htbp]
  \centering
  \includegraphics[width=1\linewidth]{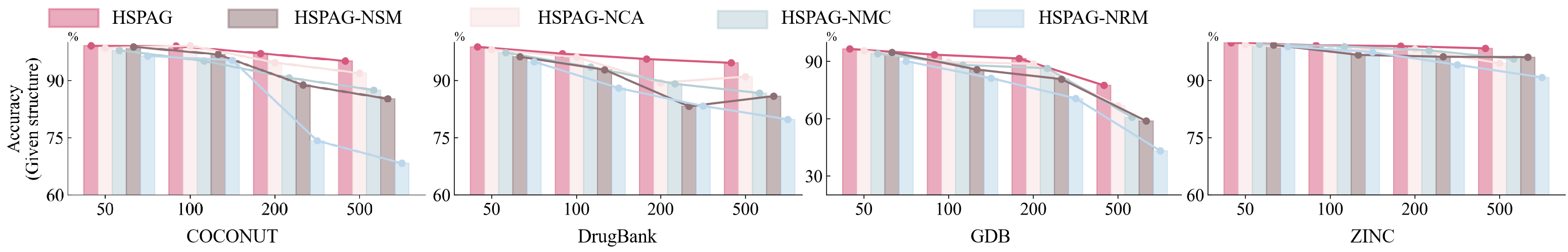}
  \caption{Ablation results on molecular retrieval task.}
  \label{ABA_r}
\end{figure}

Several modified frameworks based on \textsc{HSPAG} are introduced for ablation studies, including \textsc{HSPAG-NCA} (without clip-based augmentation), \textsc{HSPAG-NMC} (without multi-level clip alignment), \textsc{HSPAG-NSM} (without property similarity mask), and \textsc{HSPAG-NRM} (without property random mask). The results in Figure~\ref{ABA_r} demonstrate the effectiveness of each proposed component.

\clearpage
\bibliographystyle{plain}  
\bibliography{references}  

\newpage
\end{document}